\documentclass[10pt,twocolumn,letterpaper]{article}

\usepackage[pagenumbers]{wacv} 

\newcommand{\tB}[1]{\textcolor{blue}{\textbf{#1}}}

\newcommand{\Cgray}[1]{\cellcolor[HTML]{D8D6D6}} %

\usepackage{graphicx}
\usepackage{amsmath}
\usepackage{amssymb}
\usepackage{booktabs}
\usepackage{times}
\usepackage{epsfig}
\usepackage{subcaption}
\usepackage[table,xcdraw]{xcolor}
\usepackage{multirow}
\usepackage{tabularx}
\usepackage{array} 

\newcolumntype{R}{>{\centering\arraybackslash}X}

\usepackage[pagebackref,breaklinks,colorlinks]{hyperref}

\usepackage[capitalize]{cleveref}
\crefname{section}{Sec.}{Secs.}
\Crefname{section}{Section}{Sections}
\Crefname{table}{Table}{Tables}
\crefname{table}{Tab.}{Tabs.}

\begin{document}

\title{Spatial-temporal Vehicle Re-identification}

\author{Hye-Geun Kim$^1$, YouKyoung Na$^2$, Hae-Won Joe$^2$, Yong-Hyuk Moon$^{3, 4}$, Yeong-Jun Cho$^1$\\
$^1$Department of Artificial Intelligence Convergence, Chonnam National University\\
$^2$Department of Artificial Intelligence, Chonnam National University\\
$^3$Electronics and Telecommunications Research Institute\\
$^4$Department of AI, University of Science and Technology\\
{\tt\small $^1$\{hyegeunkim, yj.cho\}@jnu.ac.kr
$^2$\{me6zero, 214583\}@jnu.ac.kr 
$^{3, 4}$yhmoon@etri.re.kr }
}
\maketitle

\begin{abstract}
Vehicle re-identification~(ReID) in a large-scale camera network is important in public safety, traffic control, and security. 
However, due to the appearance ambiguities of vehicle, the previous appearance-based ReID methods often fail to track vehicle across multiple cameras. 
To overcome the challenge, we propose a spatial-temporal vehicle ReID framework that estimates reliable camera network topology based on the adaptive Parzen window method and optimally combines the appearance and spatial-temporal similarities through the fusion network.
Based on the proposed methods, we performed superior performance on the public dataset~(\texttt{VeRi776}) by 99.64\% of rank-1 accuracy.
The experimental results support that utilizing spatial and temporal information for ReID can leverage the accuracy of appearance-based methods and effectively deal with appearance ambiguities.

\end{abstract}

\section{Introduction}
\label{sec:intro}

Recently, a large number of surveillance cameras have been installed in public places for safety, traffic monitoring, and security.
However, monitoring all the cameras requires lots of human effort and resources. 
To reduce the human efforts, re-identification~(ReID) that automatically tracks targets across non-overlapping multiple cameras can be utilized.
Especially in traffic surveillance systems, vehicle ReID is essential due to the rapid flow of many vehicles on the scene.

In general, most studies have focused on targets' visual appearances to perform ReID.
For example, many studies~\cite{ahmed2015improved, chen2017person} have proposed feature learning methods to represent appearances of targets.
Similarly, metric learning methods~\cite{zheng2011person, koestinger2012large} also have been proposed to measure feature distances between query and gallery images effectively.
Recently, developments in deep learning have led to higher performance improvements by training both visual features and distance metrics~\cite{he2020fastreid,ghosh2023relation}. Those appearance-based methods are robust to target pose variations, viewpoints changes, and illumination changes.

\begin{figure}
  \centering
  \begin{subfigure}{0.85\linewidth}
    \includegraphics[width=\linewidth]{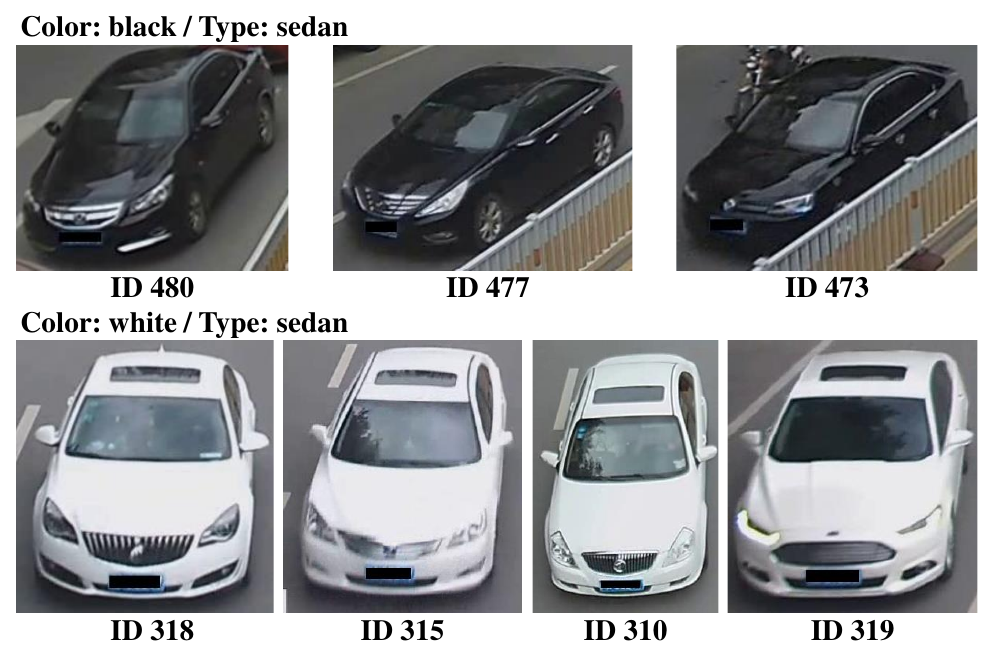}
    \caption{Appearance ambiguity problem in vehicle ReID}
    \label{subfig:sub1}
  \end{subfigure}
  \hfill
  \begin{subfigure}{0.92\linewidth}
    \includegraphics[width=\linewidth]{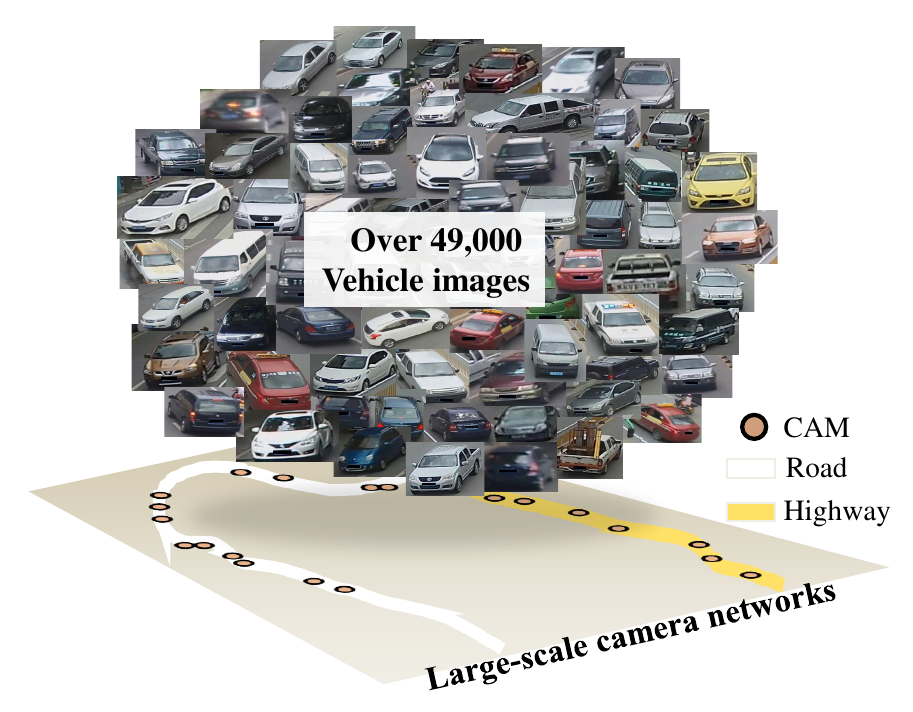}
    \caption{High computational complexity}
    \label{subfig:sub2}
  \end{subfigure}
  \caption{Challenges in vehicle re-identification}
  \label{fig:1}
\end{figure}

However, compared to person, vehicles have different characteristics leading to new challenges in re-identification.
First, vehicles with exactly the same appearances make high appearance ambiguities~(Fig.~\ref{fig:1}~(a)).
People show distinctive features, such as different faces, clothing, and body shapes, whereas many vehicles have the same appearance features due to the same model types.
Second, appearing a very large number of vehicles in multiple cameras occurs both high computational complexity and low identification performance~(Fig.~\ref{fig:1}~(b)).
Relying only on the target appearances is not effective for the vehicle ReID problem.

To alleviate the appearance ambiguity, ReID studies that use additional spatial and temporal information have been proposed~\cite{wang2019spatial, cho2019joint, shen2017learning}. 
They built a camera network topology explaining spatial and temporal relationships between cameras and utilized the topology to reduce redundant searching time ranges of queries. 
While the methods exhibit potential for improving appearance-based ReID models~\cite{he2020fastreid,ghosh2023relation}, they still have limitations. The camera network topology modeling approaches are simple, and the integration of the appearance model with spatio-temporal information lacks optimization.

%

In this work, we propose a spatial-temporal vehicle ReID framework to overcome the limitations.
The proposed ReID framework consists of two main parts: 1) camera network topology estimation, 2) fusing appearance similarity and spatial-temporal probabilities.
For the camera network topology estimation, we propose an adaptive Parzen window that is robust to outliers and sparse responses between camera pairs (Sec.~\ref{sec:proposed_1}). It can effectively handle different connection strengths of camera pairs for reliable camera network topology estimation.
After estimating the topology, we train a fusion network~(fusionNet) that can optimally combine appearance similarity and spatial-temporal probabilities (Sec.~\ref{sec:proposed_2}).
To evaluate the proposed methods, we tested \texttt{VeRi776} \cite{liu2016deep} vehicle ReID dataset.
In the experiments, we validated the effectiveness of our proposed methods.
Our methods performed the best vehicle ReID performances by 99.34\% of rank-1 accuracy and 92.40\% of mAP score compared to state-of-the-art methods.

The main contributions of this work are as follows:
\begin{itemize}
	\item We estimated a reliable camera network topology based on the proposed adaptive Parzen window. 
	\item We trained a fusion network to optimally combine two different similarities (appearance and spatial-temporal).
	\item We achieved superior performance (99.64\% of rank-1 accuracy) in the vehicle ReID task.
\end{itemize}
To the best of our knowledge, this is the first attempt to train a network for fusing appearance and spatial-temporal similarities. 
In addition, the flexibility of the proposed framework is high because any kind of appearance-based model can be used as a baseline.

\section{Related Works}
\subsection{Appearance-based ReID}
Most re-identification~(ReID) studies have focused on learning visual representations of images to distinguish their appearances. 
To this end, feature learning and metric learning methods have been widely studied.
For the feature learning method, Ahmed~\etal~\cite{ahmed2015improved} initially utilized deep convolutional neural networks~(CNN) architecture that captures local relationships between two input images based on mid-level features. Chen~\etal~\cite{chen2017person} improved the CNN-based ReID and proposed a Deep Pyramid Feature Learning (DPFL) CNN, which can learn scale-specific discriminative features.
Similarly, the studies~\cite{cheng2016person,rong2021vehicle} tried to extract robust local features.
To capture more details of appearances, Khamis~\etal~\cite{khamis2015joint} utilized attributes of targets for ReID. 
Huynh~\etal~\cite{huynh2021strong} also used multi-head with attention mechanism to improve visual representation quality. 
Recently, He~\etal~\cite{he2021transreid} proposed TransReID, which is the first attempt to utilize a transformer to learn robust features from the image patches.
For a better visual representation, Li~\etal~\cite{li2023clip} proposed CLIP-ReID, which fine-tunes the initialized visual model by the image encoder in CLIP.

For metric learning, learning the Mahalanobis distance~\cite{zheng2011person, koestinger2012large} has been widely studied. Especially, optimizing triplet loss for deep metric learning~\cite{cheng2016person,hermans2017defense, he2020fastreid} has shown superior performances in ReID tasks.
To alleviate the negative effect of intra-class variance and inter-class similarity, Bai~\etal~\cite{bai2018group} utilized inter-class triplet embedding and intra-class triplet embedding. 
Ghosh~\etal~\cite{ghosh2023relation} proposed Relation Preserving Triplet Mining (RPTM), a feature matching guided triplet mining scheme that ensures that triplets preserve natural subgroupings in object IDs. 
Meanwhile, several methods~\cite{cho2016improving, su2017pose, miao2019pose} utilized additional cues such as human pose and body parts to handle pose variations and occlusions problems in ReID.
However, relying only on appearance for ReID is still hard to alleviate appearance ambiguity.

\subsection{Spatial-temporal ReID}
To overcome the limitations of the appearance-based ReID, many studies have utilized spatial-temporal information of cameras and target objects.
In general, they employed appearance-based ReID model as their baseline, and additionally exploited the spatial and temporal information.
In spatial-temporal ReID, there are two main issues: 1) Estimating spatial-temporal information so-called camera network topology in given camera networks. 2) Utilizing the estimated camera network topology for ReID.

\begin{figure*}
	\centering
	\includegraphics[width=1\linewidth]{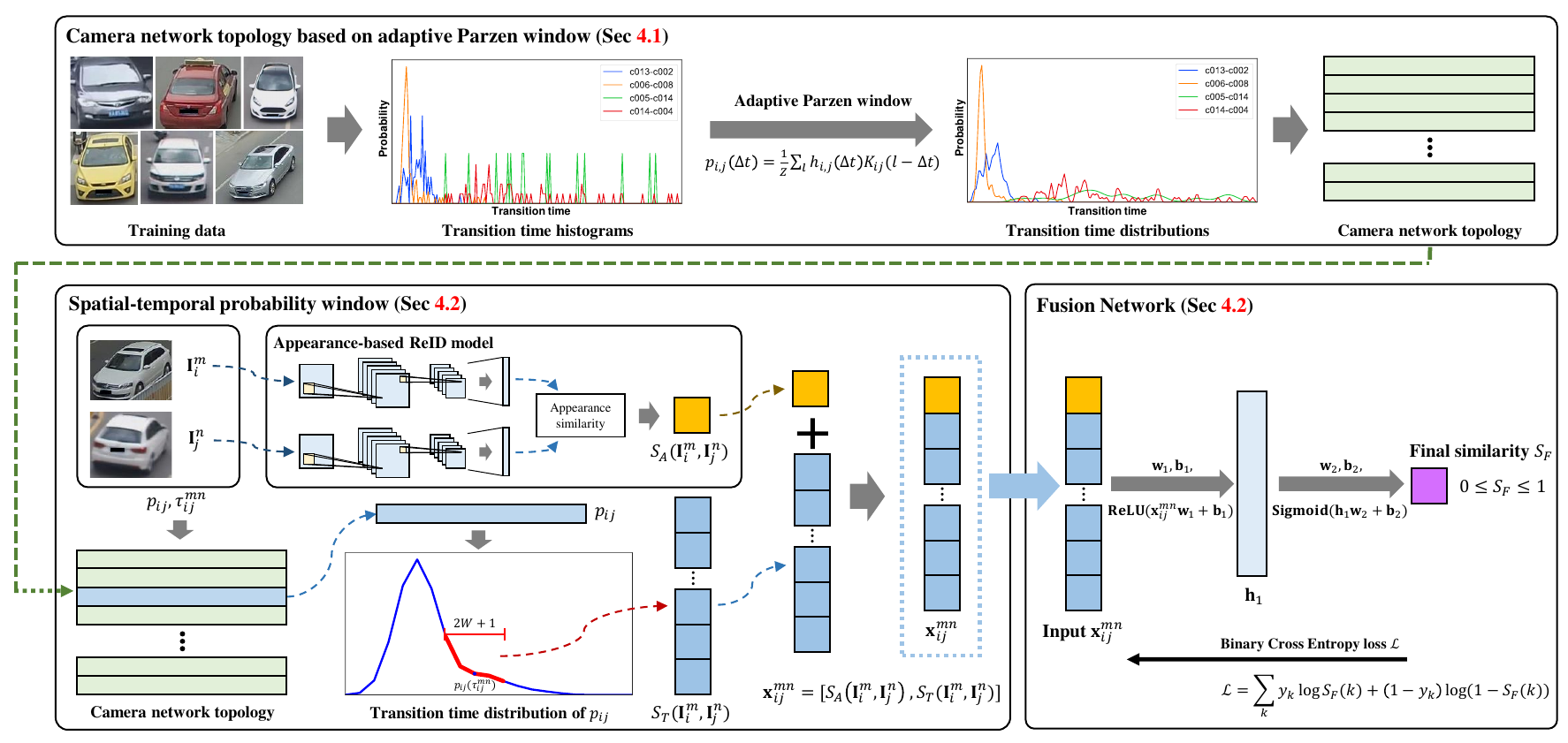}
	\caption{The overall framework for vehicle re-identification with spatial-temporal information}
	\label{fig:2}
\end{figure*} 

To estimate the camera network topology, studies~\cite{huang2016camera,cho2019joint,wang2019spatial} have tried to design accurate transition time distributions of targets~(e.g., human, vehicle).
For example, Huang~\etal~\cite{huang2016camera} utilized the Weibull distribution to formulate the pedestrian transition probability. 
Cho~\etal~\cite{cho2019joint} proposed joint estimations of camera network topology and person ReID.
Wang~\etal~\cite{wang2019spatial} proposed the Histogram-Parzen to estimate spatial-temporal probability distributions.
Also, Huang~\etal~\cite{huang2022vehicle} modeled a spatio-temporal model leveraging by vehicle pose view embedding.
Liu~\etal~\cite{liu2016deep,liu2017provid} proposed a progressive vehicle ReID that partially utilizes simple spatial-temporal information.
Similarly, the studies~\cite{lv2019vehicle,zheng2020vehiclenet, ren2021learning} estimated spatial-temporal information to filter out irrelevant gallery images.
Shen~\etal~\cite{shen2017learning} proposed a Siamese-CNN + Path-LSTM network to predict the path through visual feature information and spatial-temporal information. 

While numerous spatial-temporal ReID methods have been proposed, there are still some limitations. First, methodologies for estimating spatial-temporal models are quite simple. For example, many methods~\cite{liu2016deep,liu2017provid,lv2019vehicle} just built objects' transition time distributions based on the positive responses between cameras. However, noisy and sparse responses make the estimated distributions unreliable.
Second, usages of the spatial-temporal information are not optimized. E.g., the studies~\cite{wang2019spatial,ren2021learning,huang2022vehicle} just merged both probabilities~(i.e., appearance and spatial-temporal) with the same importance to get the joint probability.
Similarly, many methods~\cite{lv2019vehicle,zheng2020vehiclenet, ren2021learning} simply utilized spatial-temporal information to reduce the searching range or perform re-ranking the initial ReID results.

\section{Motivation and Main Ideas}
To handle the challenges in vehicle ReID, we analyzed the characteristics of vehicles in camera networks.
First, vehicles can have exactly the same appearance~(e.g., model, shape, and color) according to their model types.
Second, since vehicles can only move along roads and highways, vehicle movements across cameras can be predicted.
To sum up, compared to person ReID, vehicles show high appearance ambiguities and predictable movements.
Therefore, relying only on appearance differences between vehicles is not effective for the vehicle ReID problem.

Based on the observations of vehicles, we additionally exploit spatial and temporal relationships between cameras called a camera network topology.
As shown in Fig.~\ref{fig:2}, the proposed ReID framework consists of two main parts: 1) camera network topology estimation, 2) fusing appearance and spatial-temporal similarities.
We first build the topology based on the proposed adaptive Parzen window~(Sec.~\ref{sec:proposed_1}). 
We then train a fusion network that optimally combines visual similarity and the camera network topology information for the final ReID prediction~(Sec.~\ref{sec:proposed_2}).

\begin{figure*}
  \centering
  \begin{subfigure}{0.33\linewidth}
    \includegraphics[width=\linewidth]{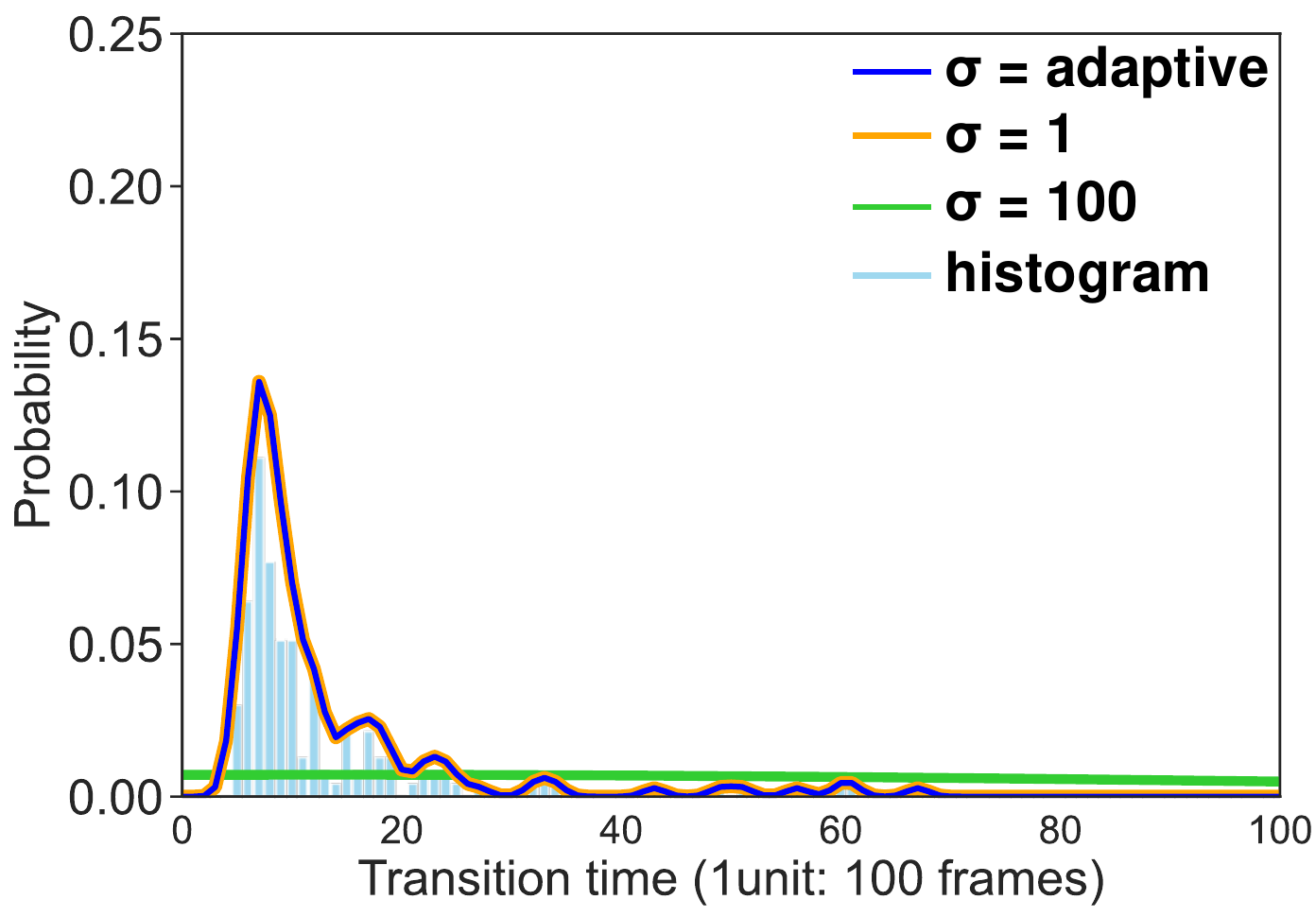}
    \caption{Vehicle pairs between cameras: 143\\ \centering (Connection strength: Strong)}
    \label{subfig:sub1}
  \end{subfigure}
  \hfill
  \begin{subfigure}{0.33\linewidth}
    \includegraphics[width=\linewidth]{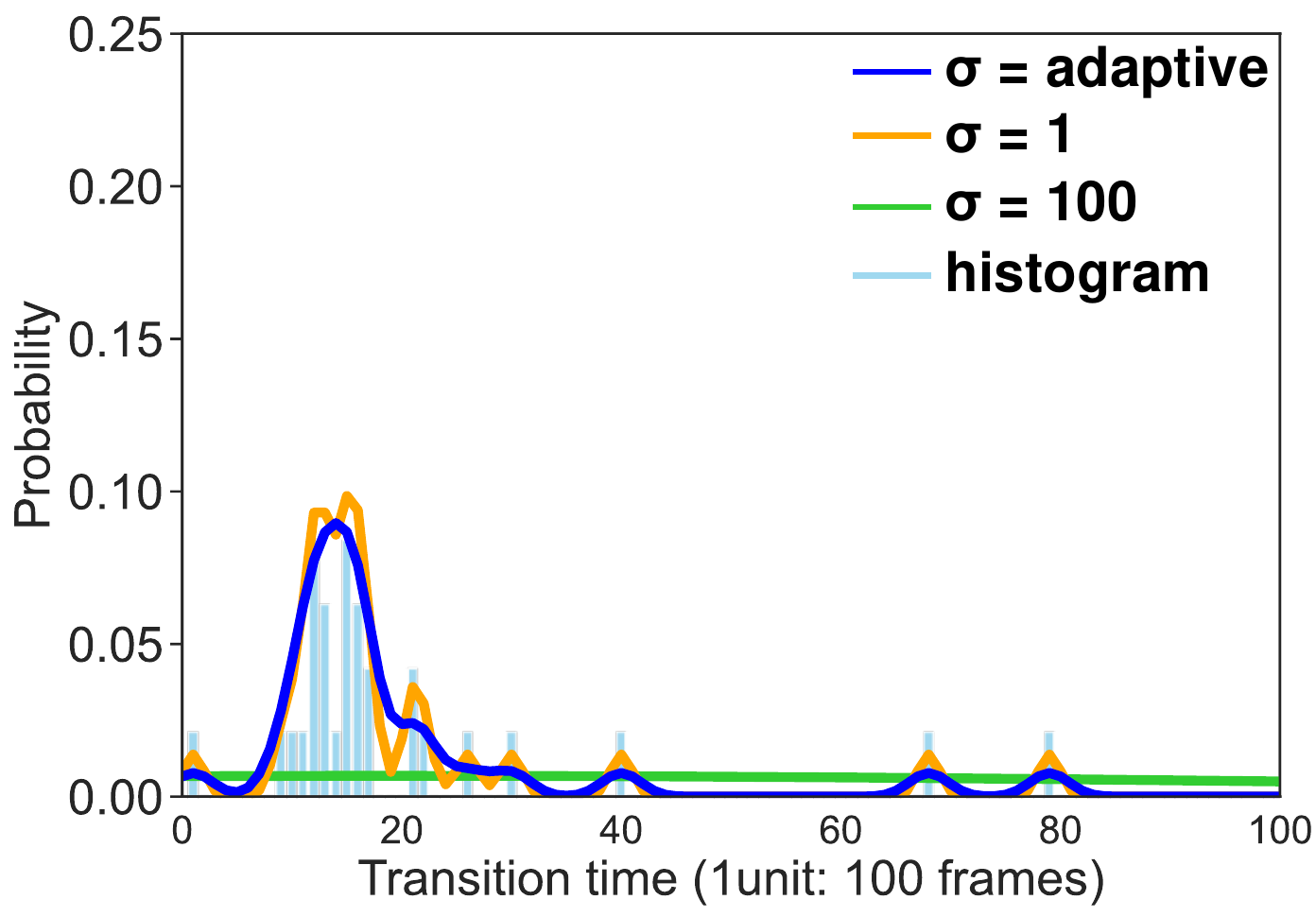}
    \caption{Vehicle pairs between cameras: 29\\ \centering (Connection strength: Normal)}
    \label{subfig:sub2}
  \end{subfigure}
  \hfill
  \begin{subfigure}{0.33\linewidth}
    \includegraphics[width=\linewidth]{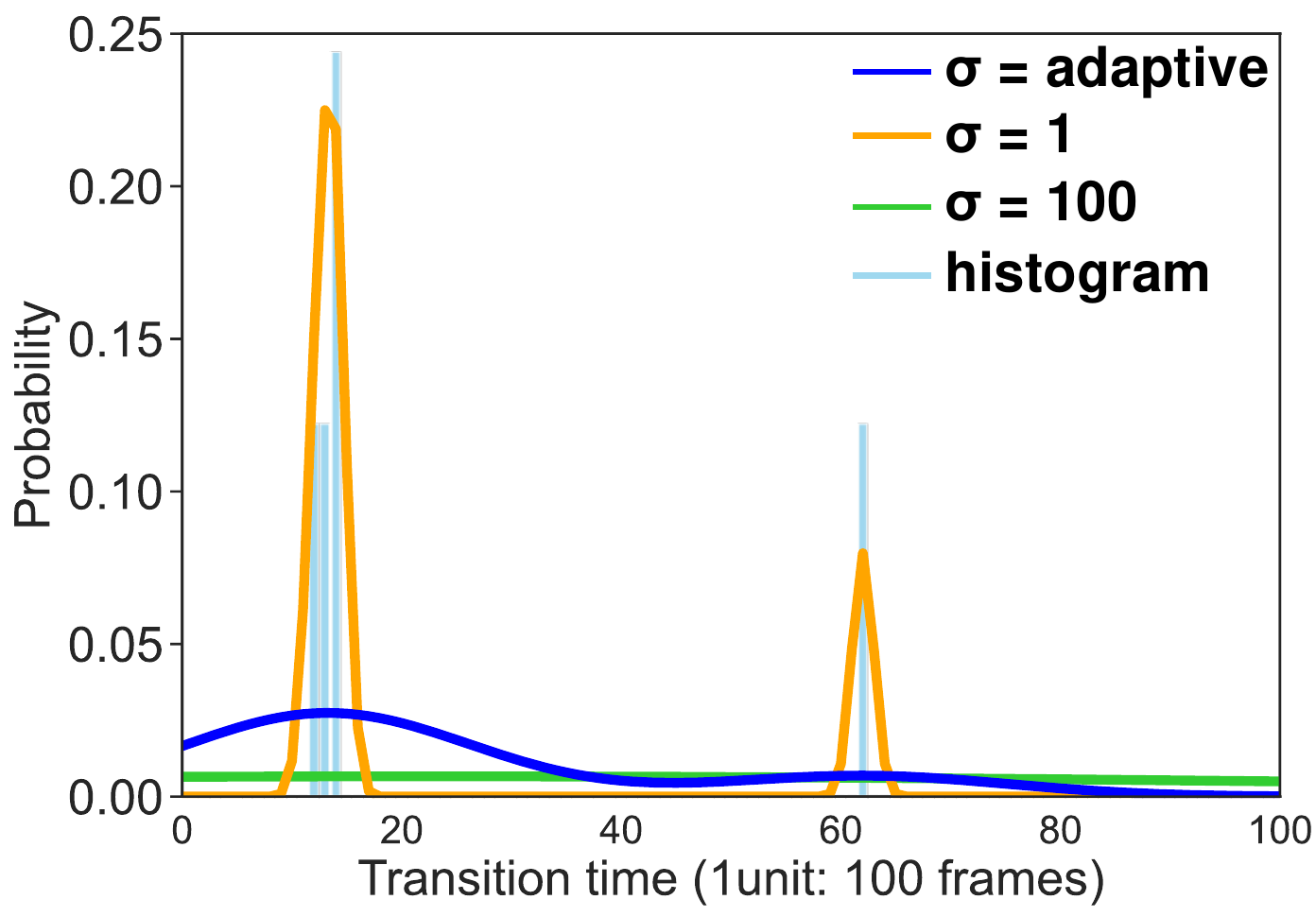}
    \caption{Vehicle pairs between cameras: 5\\ \centering (Connection strength: Weak)}
    \label{subfig:sub3}
  \end{subfigure}
  
  \caption{Examples of estimated transition time distributions between camera pairs. Each bin covers 100 frame ranges. Blue solid line (\tB{---}) is the estimated distribution~($p_{ij}$) from the histogram~($h_{ij}$) by the proposed adaptive Parzen window. Best viewed in color.}
  \label{fig:three_figures}
\end{figure*}

\section{Proposed Methods}
\label{sec:proposed}

\subsection{Adaptive Parzen window for \\ camera network topology estimation}
\label{sec:proposed_1}

Camera network topology represents spatio-temporal relationships and connections between cameras that can be represented by a graph $\mathbf{G}=(\mathbf{V},\mathbf{E})$. The vertices $\mathbf{V}$ denote cameras and the edges $\mathbf{E}$ denote distributions of object transition time.
Suppose there are $N_{cam}$ numbers of cameras in the camera networks. 
Then, the topology is represented by 
\begin{equation}
	\begin{split}
		\mathbf{V} & \in \{c_i|1\leq i\leq N_{cam}\}, \\
		\mathbf{E} & \in \{p_{ij}|1\leq i\leq N_{cam}, 1\leq j\leq N_{cam}, i\neq j\},
	\end{split} 
	\label{eq:1}
\end{equation}
where $c$ denotes a camera, $p_{ij}$ denotes an object transition time distribution between camera pairs $c_i$ and $c_j$.

To build the transition time distributions $p_{ij}$, we used positive vehicle pairs between all camera pairs in the training dataset.
Based on multiple time differences~$(\Delta t)$ of positive pairs, we can generate an initial histogram of the transition time $h_{ij}(\Delta t)$ as depicted in cyan lines (\textbf{\textcolor{cyan}{---}}) in Fig.~\ref{fig:three_figures}.
Cho~\etal~\cite{cho2019joint} proposed connectivity checking criteria whether a pair of cameras is connected or not by fitting a Gaussian model to the histogram $h_{ij}(\Delta t)$. However, this parametric method followed strong assumptions and it is hard to handle outliers and sparsity of the histogram. 
	
Inspired by~\cite{wang2019spatial}, a Parzen window method can be applied to the initial histograms, and we can estimate probability density function~(PDF) of vehicle transition time in a non-parametric manner as follows
\begin{equation}
	p_{ij}\left(\Delta t\right) = \frac{1}{Z} \sum_{\tau} h_{ij}\left(\Delta t\right)K\left(\tau-\Delta t\right),
	\label{eq:2}
\end{equation}
where $Z=\sum_{\Delta t} h_{ij}(\Delta t)$ is a normalized factor and $K(\cdot)$ is a kernel function.
For the kernel $K$, the work~\cite{wang2019spatial} used Gaussian function as
\begin{equation}
	K(x) = \frac{1}{\sqrt{2\pi}\sigma}\exp\left({\frac{-x^2}{2\sigma^2}}\right).
	\label{eq:3}
\end{equation}

While the Parzen window method efficiently estimates continuous probability density functions from discrete histograms, employing a single kernel across diverse histograms from various camera pairs is not reasonable.
The strength of the spatio-temporal connection between cameras can be determined by how many vehicles pass through those cameras during a certain period of time~\cite{cho2019joint}.
For example, few positive pairs between two cameras indicate a weak connectivity between them.
Nevertheless, the Parzen window method extremely enlarges those small responses with a small $\sigma$ value as depicted in the orange line (\textbf{\textcolor{orange}{---}}) in Fig~\ref{subfig:sub3}.
In that case, it is better to use a large $\sigma$ value to avoid overfitting the distribution for noise and outliers.

On the other hand, if there are lots of positive pairs between cameras, then the connectivity should be strong. 
However, with the large $\sigma$, the resulting distribution becomes uniform, thereby failing to capture any meaningful spatial and temporal relationships between the cameras (green line (\textbf{\textcolor{green}{---}}) in Fig.~\ref{subfig:sub1}).
In that case, it is better to use a relatively small $\sigma$ value to reflect temporal information between cameras.
Thus, selecting the proper $\sigma$ value is important to the estimated distribution $(p_{ij})$ quality.

To overcome the limitation of the original Parzen window method, we newly propose adaptive Parzen window by setting different $\sigma_{ij}$ values for different camera pairs~$(c_i,c_j)$.
To this end, we designed an adaptive standard deviation according to the different strength of the camera connectivity as follow
\begin{equation}
\sigma_{ij} = \max \left(  \alpha \exp\left({\frac{-N_{ij}}{\beta}}\right), 1\right),
\label{eq:4}
\end{equation}
where $N_{ij}$ is the number of positive vehicle pairs between camera $c_i$ and $c_j$. $\alpha$ is a scale factor that determines the maximum range of $\sigma_{ij}$. $\beta$ is a smoothness factor that adjusts sensitivity of $\sigma_{ij}$. 
The minimum value of $\sigma_{ij}$ cannot be less than 1 unit of the histogram. Then, the values of $\sigma_{ij}$ lie on $\left[1, \alpha\right]$.

By considering camera indexes, Eq.~\ref{eq:2} and Eq.~\ref{eq:3} reformulated as
\begin{equation}
	p_{ij}\left(\Delta t\right) = \frac{1}{Z} \sum_{\tau} h_{ij}\left(\Delta t\right)K_{ij}\left(\tau-\Delta t\right),
\end{equation}
\begin{equation}
	K_{ij}(x) = \frac{1}{\sqrt{2\pi}\sigma_{ij}}\exp\left({\frac{-x^2}{2\sigma_{ij}^2}}\right).
\end{equation}
As a result, we can estimate reliable distributions~($p_{ij}$) from the initial discrete histograms~($h_{ij}$) by considering the connectivity between cameras. The blue lines (\tB{---}) in Fig.~\ref{fig:three_figures} are our results based on the adaptive Parzen window.

	\subsection{Fusion Network}
	\label{sec:proposed_2}

To estimate appearance similarities between images, the proposed ReID framework can employ any appearance-based ReID method as its baseline. 
Images from each camera~$(c_i,c_j)$ are denoted by $\mathbf{I}^{m}_i,\mathbf{I}^{n}_j$, where $m,n$ are indexes of the images.
Then, the appearance-based ReID methods estimate the visual similarity between two images as $S_A(\mathbf{I}^{m}_i,\mathbf{I}^{n}_j)$ that lies on $\left[0, 1\right]$.
Note that the proposed framework does not depend on the types of appearance-based models.

In order to perform spatial-temporal ReID, Cho~\etal~\cite{cho2019joint} used camera network topology only to restrict the searching range of the gallery.
It is effective to reduce the complexity of ReID, but the spatial-temporal probability has no effect on the final similarity.
On the other hand, previous studies~\cite{wang2019spatial,ren2021learning,huang2022vehicle} just merged both probabilities~(i.e., appearance and spatial-temporal) with the same importance to get the joint probability.
However, they neglected two points. First, the domain of each probability is not the same. Second, both appearance and spatial-temporal probabilities can be imperfect. Therefore, it is not reasonable to merge those probabilities half and half.
	
In this work, we optimally combines visual similarities $S_A(\mathbf{I}^{m}_i,\mathbf{I}^{n}_j)$ and estimated spatial-temporal distributions $p_{ij}\left(\Delta t\right)$ based on the fusion network.
An input vector of the network for two images $(\mathbf{I}^{m}_i,\mathbf{I}^{n}_j)$ in the camera pair $(i,j)$ can be represented by
\begin{equation}
    \mathbf{x}^{mn}_{ij} = [S_A(\mathbf{I}^{m}_i,\mathbf{I}^{n}_j), S_T(\mathbf{I}^{m}_i,\mathbf{I}^{n}_j)],
    \label{eq:6}
\end{equation}
where $S_A$ is an appearance similarity, and $S_T$ is a spatial-temporal vector.
$S_T$ vector between images are defined by		
\begin{equation}
    \begin{split}
        S_T(\mathbf{I}^{m}_i,\mathbf{I}^{n}_j) = \biggl[ & p_{ij}\left(\tau^{mn}_{ij}-W\right), ... , \\
        & p_{ij}\left(\tau^{mn}_{ij}-1\right),p_{ij}\left(\tau^{mn}_{ij}\right),p_{ij}\left(\tau^{mn}_{ij}+1\right), \\
        & ...,p_{ij}\left(\tau^{mn}_{ij}+W\right) \biggr],
    \end{split}
    \label{eq:7}
\end{equation}
where $\tau^{mn}_{ij}$ is the time difference between two images $\mathbf{I}^{m}_i$ and $\mathbf{I}^{n}_j$.
$W$ is the size of a time window. According to the $W$, the range of the $S_T$ vector is determined around distributions of $p_{ij}\left(\tau^{mn}_{ij}\right)$. E.g. when we set $W=0$, the $S_T$ becomes a scalar value as $S_T(\mathbf{I}^{m}_i,\mathbf{I}^{n}_j) = p_{ij}\left(\tau^{mn}_{ij}\right)$.
When we set $W>0$, the $S_T$ vector has a $2W+1$ dimensional vector.
By adjusting the value of $W$, we can determine how much spatio-temporal information to provide for the fusion network.
    Then, the dimension of the input vector $\mathbf{x}^{mn}_{ij}$ for the fusionNet is $2W+2$.

We designed the fusion network based on simple Multi-layer Perception~(MLP). We empirically found that the fusion network does not require sophisticated deep neural network structure to estimate the final similarity.
The network has one hidden layer with several nodes and a one-dimensional output layer, as shown in Fig.~\ref{fig:2}.
For the activation function, we used Rectified Linear Unit~(ReLU) for nodes in the hidden layer,
and sigmoid function for the output node. Then, the final output of the fusion network $S_F(\mathbf{I}^{m}_i,\mathbf{I}^{n}_j)$ lies on $\left[0,1\right]$.
To train the network, we optimized binary cross entropy loss defined by
\begin{equation}
    \mathcal{L} =\sum_k{y_k \log{S_F\left(k\right)}+\left(1-y_k\right)\log(1-{S_F\left(k\right))}},	
\end{equation}
where $k$ is an index of training image pair, $y_k\in[0,1]$ is the ground-truth of the $k$-th image pair.

\section{Experimental Results}
\label{sec:exp}

\subsection{Dataset and Settings}
\label{sec:exp_1}

For experiments, we used the \texttt{VeRi776}~\cite{liu2016deep} vehicle re-identification~(ReID) dataset.
It contains over 49,000 images of 776 different vehicles captured by 20 non-overlapping synchronized cameras $(N_{cam} = 20)$. 
Each image contains vehicle ids, timestamps~(frame No.), and camera ids.
We used 37,778 training images from 576 vehicle identities to train the appearance-based ReID model and the camera network topology $\mathbf{G}=(\mathbf{V},\mathbf{E})$.
For the appearance-based ReID model, we trained FastReID~\cite{he2020fastreid} to extract the appearance similarity. 
ResNet-50~\cite{he2016deep} is set to its backbone network structure and the training parameters are as follows: 
epoch -- 60, batch size -- 64. 
Note that we can use other state-of-the-art appearance-based ReID models for our framework.
The estimated camera network topology consists of 400 object transition time distributions $(p_{ij})$.
Among them, 380 distributions are between different camera pairs (i.e., $p_{ij}$, where $c_i \neq c_j$), and
the other 20 distributions are the distributions between themselves (i.e., $p_{ij}$, where $c_i = c_j$).
Each distribution has 300 bins, and a bin covers 100 frame ranges.
All the distributions were estimated based on the proposed adaptive Parzen window.

The proposed fusion network~(fusionNet) has a single hidden layer, and we designed the number of nodes in the hidden layer to be around 65\% of the size of the input vector by round$(2(2W+2)/3+1)$.
The training parameters of the network are as follows: epoch -- 100, batch size -- 128, learning rate -- 0.001, optimizer -- Adam. 
To train fusionNet and evaluate the ReID performance, we utilized 1,678 query images and 11,579 gallery images from 200 vehicle identities.
For fair comparison, a 5-fold cross-validation was conducted, and we evaluated Top-\textit{k} accuracy $(k=1, \ 5)$ and mean Average Precision~(mAP) scores.

\begin{table}[t]
	\small
	\centering
	\begin{tabularx}{\columnwidth}{c|R|R|R}
		\hline
		\textbf{Methods} & \textbf{Rank-1} & \textbf{Rank-5} & \textbf{mAP} \\ \hline
		\textit{Only appearance}~\cite{he2020fastreid} & 96.96 & 98.45 & 81.91 \\ \hline
		\textit{Without fusionNet}~\cite{wang2019spatial} & 95.77 & 97.74 & 85.47 \\ \hline\hline
		fusionNet $(W=0)$   & 98.21 & 99.70 &  89.66 \\ \hline
		fusionNet $(W=2)$   & 99.17 & 99.76 & 92.42 \\ \hline
		fusionNet $(W=4)$   & 99.23 & 99.70 & 91.83 \\ \hline
		fusionNet $(W=6)$   & 99.40 & 99.64 & 92.70 \\ \hline
		fusionNet $(W=8)$   & 99.34 & 99.70 & 92.47 \\ \hline
        fusionNet $(W=10)$   & \textbf{99.64} & \textbf{99.82} & \textbf{92.79} \\ \hline
        fusionNet $(W=12)$   & 99.46 & \textbf{99.82} & 92.76 \\ \hline
	\end{tabularx}
	\caption {{ReID performances according to $W$ values in fusionNet. The method \textit{Only appearance} does not exploit spatial-temporal information $S_T$. The method \textit{without fusionNet} estimates the final similarity  by $S_F=S_A(\textbf{I}^m_i,\textbf{I}^n_j) \cdot p_{ij}\left(\tau^{mn}_{ij}\right)$ as in~\cite{wang2019spatial}.}}
	\label{tab:1}
\end{table}

\subsection{Effects of fusionNet}
\label{sec:exp_3}

The proposed fusionNet that combines appearance similarity~$S_A$ and spatial-temporal distributions~$S_T$ can improve the ReID performance.
To compare the performance, we also evaluated two baseline methods (1.\textit{Only appearance}, 2.\textit{Without fusionNet}). \textit{Only appearance} based on FastReID~\cite{he2020fastreid} performed ReID only using appearance similarities~$(S_A)$ between images. Another baseline \textit{Without fusionNet} estimated appearance similarity~$(S_A)$ by FastReID~\cite{he2020fastreid} and simply combined spatial-temporal similarity by $S_F=S_A(\textbf{I}^m_i,\textbf{I}^n_j) \cdot p_{ij}\left(\tau^{mn}_{ij}\right)$ as in~\cite{wang2019spatial}.
For the fusionNet, we tested various $W$ values to determine the spatial-temporal vector $S_T$.

As shown in Table~\ref{tab:1}, the appearance-based ReID~\cite{he2020fastreid} achieved rank-1 accuracy of 96.96\% and 81.91\% of mAP score. That is quite high performance in vehicle ReID, but there is still room for improvement.
Using additional spatial-temporal similarity without fusionNet, it could improve the mAP score of appearance-based ReID~\cite{he2020fastreid} by 3.56\%.
However, combining strategy of two different similarities~$S_A(\textbf{I}^m_i,\textbf{I}^n_j)$, $p_{ij}\left(\tau^{mn}_{ij}\right)$ is not optimized yet.
Rank-1 and rank-5 performances were rather degraded after using spatial-temporal information.

\begin{figure}
	\centering
	\includegraphics[width=1\linewidth]{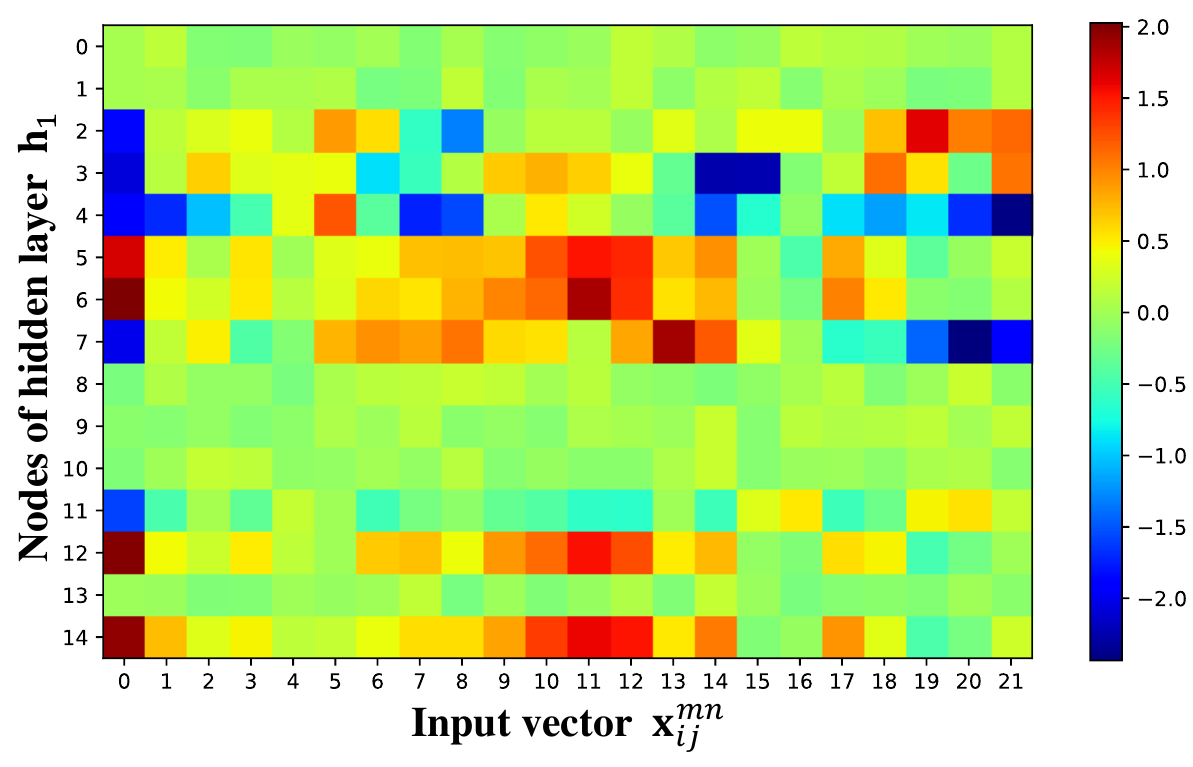}
	\caption{A visualization of the trained weight vector~($\mathbf{w}_1$) between input and hidden layer~($\mathbf{h}_1$)}
	\label{fig:4}
\end{figure}

Based on the proposed fusionNet, we can further improve the vehicle ReID performance.
When the $W$ was set to 10, fusionNet achieved the best performance by rank-1 accuracy of 99.64\%, rank-5 accuracy of 99.82\% and 92.79\% of mAP score.
Except for $W=0$, fusionNet outperformed other methods in both evaluation metrics (rank-1,-5, mAP). 
This result supports that the proposed fusionNet can optimally combine different types of information such as appearance similarity and spatial-temporal information.

Figure~\ref{fig:4} shows a visualization of the trained weight vector~($\mathbf{w}_1$) between input and hidden layer~($\mathbf{h}_1$). The row numbers~(0--14) denote the index of the nodes in $\mathbf{h}_1$, and column numbers~(0--22) denote the index of the input vector $\mathbf{x}^{mn}_{ij}$.
The first column~(0 index) shows the weights for appearance similarity~($S_A$). 
As we can see, the magnitudes of the weights are relatively bigger than those of other weights. 
It means that the fusion network properly trained the importance of the appearance similarity~($S_A$).
The other columns~(from 0 to 21 index) are the weights for spatial-temporal distribution~($S_T$).
Interestingly, the weights from the 10th to 12th columns showed large magnitudes.
It implies that the spatial-temporal information around the time difference ($\tau^{mn}_{ij}$) between two images plays a key role in ReID.

\begin{figure}
	\centering
	\includegraphics[width=0.9
 \linewidth]{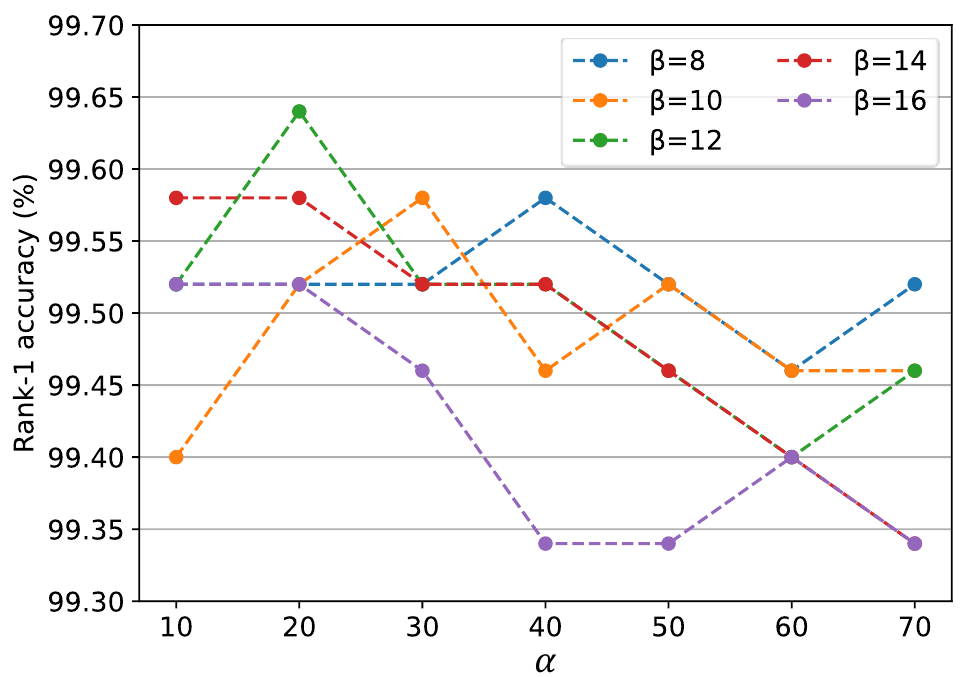}
	\caption{Performances according to factors $\alpha$ and $\beta$ of the adaptive Parzen window}
	\label{fig:5}
\end{figure}

\subsection{Effects of adaptive Parzen window}
\label{sec:exp_2}

For the proposed adaptive Parzen window method, we tested various factors such as a scale factor $\alpha$ and smoothness factor $\beta$ in Eq.~\ref{eq:4}. In all experiments, we set $W=10$ for the fusionNet. Figure~\ref{fig:5} illustrates rank-1 performances according to the factors $\alpha$ and $\beta$.
When the scale factor $\alpha=20$ and the smoothness factor $\beta=12$, the proposed framework achieved the best performance at rank-1 accuracy. Note that the rank-1 performance does not fluctuate significantly due to the factor values (min: 99.34, max: 99.64).

We then compared ReID performances of the fixed $\sigma$ and adaptive $\sigma$.
We tested the fixed $\sigma$ by 1, 10, and 300, and the adaptive $\sigma$ as in Eq.~\ref{eq:2} as shown in Table~\ref{tab:2}.
The fixed $\sigma=300$ showed the lowest performance, because too large $\sigma$ leads smoothed distributions that is close to uniform distribution, and reduces the impact of spatial-temporary information.
$\sigma=1$ and $\sigma=10$ performed relatively well, achieving higher than 99\% rank-1 accuracy.
Compared to using fixed $\sigma$ values, the proposed adaptive Parzen window with the adaptive $\sigma$ achieved the best performance in both evaluation metrics by the 99.64\% rank-1 accuracy, 99.82\% rank-5 accuracy and 92.79\% mAP score.
As we explained in Sec.~\ref{sec:proposed_1} and Fig~\ref{fig:three_figures}, a fixed value of $\sigma$ is hard to handle the various types of initial histograms $h_{ij}$.
This result implies that setting different $\sigma_{ij}$ values by considering the different connection strengths of camera pairs is effective and improves the ReID performance.

\begin{table}[]
\centering
\begin{tabularx}{\columnwidth}{l|R|R|R}
\hline
\textbf{Methods} & \textbf{Rank-1} & \textbf{Rank-5} &  \textbf{mAP} \\ \hline
fixed $\sigma=1$         & 99.46 & 99.70 & $\mathbf{92.79}$ \\
fixed $\sigma=10$        & 99.52 & $\mathbf{99.82}$ & 92.61 \\
fixed $\sigma=100$       & 99.52  & $\mathbf{99.82}$ & 89.97 \\
fixed $\sigma=300$       & 99.17 & 99.70  & 87.98 \\
(Ours) adaptive $\sigma$ & $\mathbf{99.64}$ & $\mathbf{99.82}$ & $\mathbf{92.79}$ \\ \hline
\end{tabularx}
\caption{ReID performances according to the values of $\sigma$ for Parzen window method}
\label{tab:2}
\end{table}

\subsection{Comparison with state-of-the-art methods}
\label{sec:exp_4}
In this section, we compare the proposed method with state-of-the-art re-identification methods using a \texttt{VeRi776}~\cite{liu2016deep} vehicle ReID dataset.
The methods are mainly categorized into two approaches: 1) only appearance-based, 2) using additional spatial-temporal information~(marked by $\dagger$).
The methods marked by $\star$ performed re-ranking post-processing for final ReID results.
Vehicle ReID results are summarized in Table.~\ref{tab:3}, and the list is sorted by rank-1 performance.

Compared to appearance-based approaches, fewer spatial-temporal approaches have been proposed.
Although the spatial-temporal approaches~\cite{liu2016deep,shen2017learning,liu2017provid} have improved their baseline methods, the performances are relatively low than other state-of-the-art methods. 
That is because their appearance models were slightly old-fashioned methods such as SIFT, Bag-of-words, and Siamese-CNN for ReID.
Furthermore, the methods did not provide a direct estimation of the camera network topology, nor did they optimize the utilization of spatial-temporal information.
For the fair comparison, we implemented a spatial-temporal vehicle ReID model~(\textit{Without fusionNet}) by combining the methods FastReID~\cite{he2020fastreid} and Wang's~\cite{wang2019spatial} as we explained in Sec.~\ref{sec:exp_3}. Although it has the same appearance-based model as our method, our rank1 performance is 3.87\% better.

\begin{table}[]
	\centering
	\fontsize{8.1}{12}\selectfont
	\begin{tabularx}{\columnwidth}{l|R|R|R}
		\hline
		\textbf{Models}  & \textbf{Rank-1} & \textbf{Rank-5} & \textbf{mAP} \\ \hline
		$\dagger$FACT+Plate-SNN+STR~\cite{liu2016deep} & 61.44 & 78.78 & 27.77 \\
		$\dagger$Siamese-CNN+Path-LSTM~\cite{shen2017learning} & 83.49 & 90.04 & 58.27 \\
		$\dagger$PROVID~\cite{liu2017provid}           & 81.56 & 95.11 & 53.42 \\ 
		$\dagger$KPGST~\cite{huang2022vehicle} & 92.35 &  93.92 & 68.73 \\
		$\dagger$FastReID~\cite{he2020fastreid} + Wang's~\cite{wang2019spatial} & 95.77 & 97.74 & 85.47 \\ \hline
		$\star$GAN+LSRO~\cite{wu2018joint}            & 88.62 & 94.52 & 64.78 \\
		$\star$AAVER~\cite{khorramshahi2019dual}      & 90.17 & 94.34 & 66.35 \\
		PAMTRI~\cite{tang2019pamtri}           & 92.86 & 96.97 & 71.88 \\
		SPAN~\cite{chen2020orientation}        & 94.00 & 97.60 & 68.90 \\
		CAL~\cite{rao2021counterfactual}       & 95.40 & 97.90 & 74.30 \\
		PVEN~\cite{meng2020parsing}            & 95.60 & 98.40 & 79.50 \\
		TBE~\cite{sun2021tbe}                  & 96.00 & 98.50 & 79.50 \\
		TransReID~\cite{he2021transreid}       & 96.90 & - & 80.60 \\ 
		$\star$VehicleNet~\cite{zheng2020vehiclenet}  & 96.78 & - & 83.41 \\
		$\star$SAVER~\cite{khorramshahi2020devil}     & 96.90 & 97.70 & 82.00 \\
		$\star$DMT~\cite{he2020multi}                 & 96.90 & - & 82.00 \\
		FastReID~\cite{he2020fastreid} & 96.96 & 98.45 & 81.91 \\
        CLIP-ReID~\cite{li2023clip}    & 97.30 & - & 84.50 \\
		$\star$Strong Baseline~\cite{huynh2021strong} & 97.00 & - & 87.10 \\
		$\star$RPTM~\cite{ghosh2023relation}          & 97.30 & 98.40 & 88.00 \\ \hline
		\textbf{Ours}                          & $\mathbf{99.64}$ & $\mathbf{99.82}$ & $\mathbf{92.79}$ \\ \hline
	\end{tabularx}
\caption {Performance comparisons on~\texttt{VeRi776}~\cite{liu2016deep} data. $\dagger, \star$ denote spatial-temporal approach and re-ranking, respectively.}
\label{tab:3}
\end{table}

As the deep learning model developments, many appearance-based methods have improved ReID performances without spatial-temporal information.
For example, FastReID~\cite{he2020fastreid} which is our baseline appearance model achieved 96.96\% rank-1 accuracy and 81.91\% mAP score. Especially, RPTM~\cite{ghosh2023relation} used GMS~\cite{bian2017gms} feature matcher and employed ResNet-101~\cite{he2016deep} structure showed the best performance by 97.30\% of rank-1 accuracy, 88.00\% of mAP score.
Although our method used lightweight structure~(ResNet-50) for the appearance model, we outperformed rank-1 accuracy of 2.34\%, rank-5 accuracy of 1.37\% and mAP score of 4.79\% over the best score achieved by SOTA methods.
In addition, the proposed method did not perform any re-ranking processes for the post-processing.

\begin{figure*}
	\centering
	\includegraphics[width=1\linewidth]{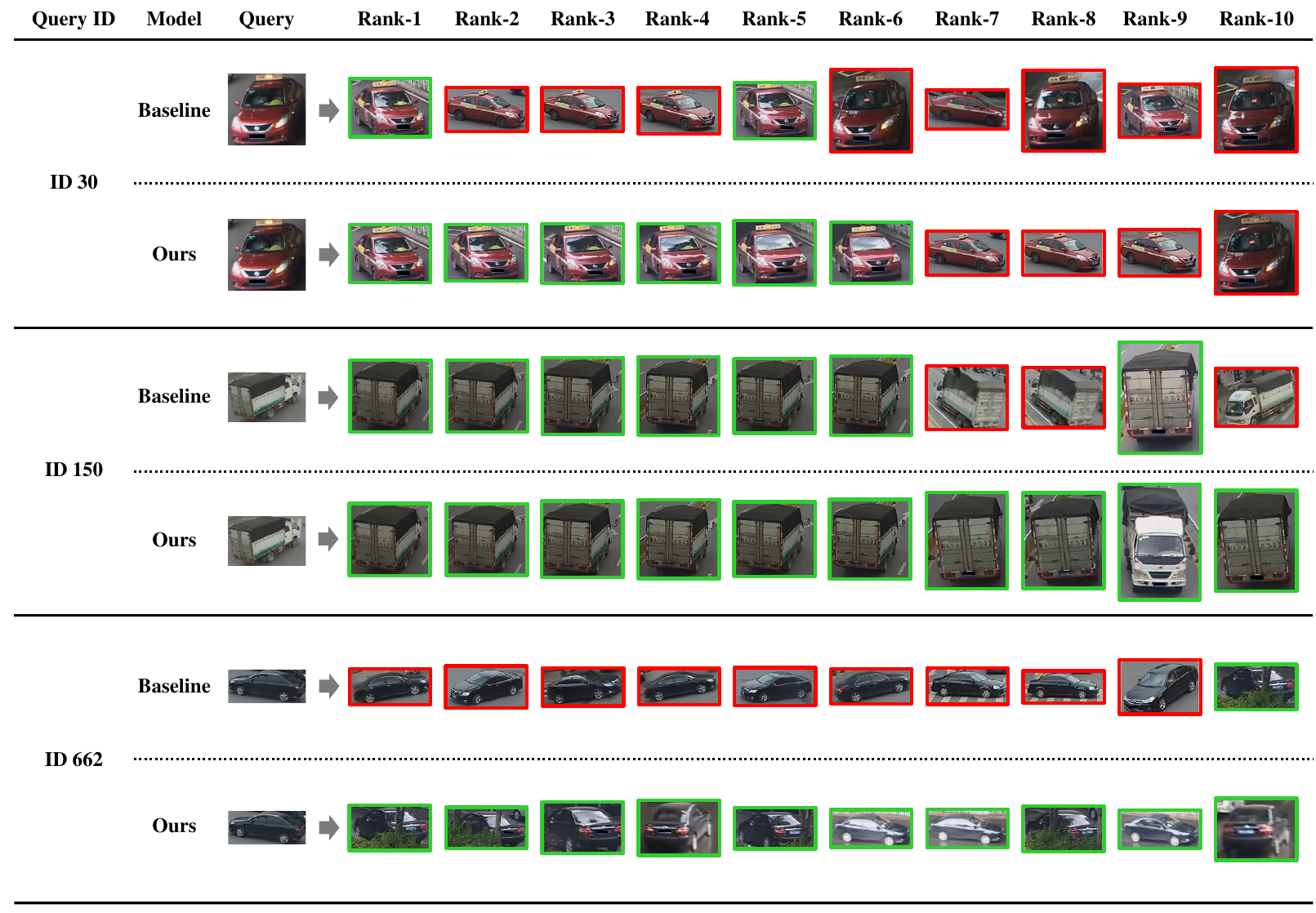}
	\caption{Qualitative vehicle ReID results of baseline method and ours. The baseline method~\cite{he2020fastreid} is an appearance-based method. Green and red boxes denote the true and false matching, respectively. Compared to baseline method, the proposed method perfectly matches the true positive pairs despite of similar appearances. Best viewed in color.}
	\label{fig:6}
\end{figure*}

Figure.~\ref{fig:6} illustrates qualitative vehicle ReID results. 
We compared our method with a baseline method (FastReID~\cite{he2020fastreid}) which used only appearance information. The baseline method occurs numerous false matches, where the appearances closely resemble that of the query image.
In particular, the baseline model rarely matched correct images of the 662-th query image due to many similar cars. 
On the other hand, our method perfectly matched correct images at rank-1 to rank-10 under those challenging query and gallery pairs.
These results support that the proposed spatial-temporal vehicle ReID with fusionNet can effectively handle the appearance ambiguity problems, and overcome the limitations of the previous ReID methods.
It is worth highlighting that the proposed vehicle ReID is simple but effective. Furthermore, it has high compatibility with various baseline models.

\section{Conclusion}
In this work, we proposed a vehicle ReID framework that can estimate camera network topology and
combines appearance and spatial-temporal similarities to alleviate appearance ambiguities.
To this end, we proposed an adaptive Parzen window for reliable topology estimation and a fusion network~(fusionNet) for optimal similarity aggregation.
The proposed framework achieved superior performance for vehicle ReID with the rank-1 accuracy of 99.64\% and mAP score of 92.79\% on \texttt{VeRi776} data.
The results support that utilizing spatial and temporal information for ReID can leverage the accuracy of appearance-based methods and effectively deal with appearance ambiguities.

We confirmed the potential for the proposed framework and there are a few areas that could be explored for future work.
First, we will utilize another appearance-based method as the baseline of the framework.
Since satisfying results have been obtained by the most common baseline~(ResNet structure with triplet loss optimization), we expect that our framework can improve other baselines as well.
Second, we will extend our framework to other domains such as person ReID.


{\small
\bibliographystyle{ieee_fullname}
\bibliography{egbib}
}

\end{document}